\pdfoutput=1

\documentclass[11pt,a4paper]{article}
\usepackage{authblk}
\usepackage[final]{naaclhlt2021}
\usepackage[utf8]{inputenc}
\usepackage[T1]{fontenc}
\usepackage{times}
\usepackage{latexsym}
\usepackage{cleveref}
\usepackage{enumitem}
\usepackage{graphicx}
\usepackage{booktabs}
\usepackage{makecell}
\usepackage{url}
\usepackage{multirow}
\usepackage[normalem]{ulem} 
\usepackage{xcolor}

\newcommand{\ptakopet}{Ptakopět}

\definecolor{darkgreen}{rgb}{0.2,0.7,0.2}
\definecolor{darkred}{rgb}{0.7,0.2,0.2}

\newcommand*{\SuperScriptSameStyle}[1]{%
  \ensuremath{%
    \mathchoice
      {{}^{\displaystyle #1}}%
      {{}^{\textstyle #1}}%
      {{}^{\scriptstyle #1}}%
      {{}^{\scriptscriptstyle #1}}%
  }%
}

\newcommand*{\oneSmark}{\SuperScriptSameStyle{*}}
\newcommand*{\twoSmark}{\SuperScriptSameStyle{\diamond}}
\newcommand*{\threeSmark}{\SuperScriptSameStyle{\dagger}}

\newcommand*{\oneS}{\rlap{\oneSmark}}
\newcommand*{\twoS}{\rlap{\twoSmark}}
\newcommand*{\threeS}{\rlap{\threeSmark}}

\title{Backtranslation Feedback Improves User Confidence in MT, Not Quality}

\author[1]{Vilém Zouhar}
\author[1]{Michal Novák}
\author[1]{Matúš Žilinec}
\author[1]{Ondřej Bojar}
\author[2]{Mateo Obregón}
\author[2]{\\Robin L. Hill}
\author[3,4]{Frédéric Blain}
\author[3]{Marina Fomicheva}
\author[3]{Lucia Specia}
\author[5]{Lisa Yankovskaya}

\makeatletter
\renewcommand\AB@affilsepx{, \protect\Affilfont}
\makeatother
\affil[1]{Charles University, Faculty of Mathematics and Physics, Czech Republic}
\makeatletter
\renewcommand\AB@affilsepx{\\\protect\Affilfont}
\makeatother
\affil[5]{University of Tartu, Estonia}
\makeatletter
\renewcommand\AB@affilsepx{, \protect\Affilfont}
\makeatother
\affil[2]{University of Edinburgh}
\affil[3]{University of Sheffield}
\makeatletter
\renewcommand\AB@affilsepx{\\\protect\Affilfont}
\makeatother
\affil[4]{University of Wolverhampton, United Kingdom}

\makeatletter
\renewcommand\AB@affilsepx{ \protect\Affilfont}
\makeatother
\affil[ ]{\small\tt\{zouhar,mnovak,zilinec,bojar\}@ufal.cuni.cz}
\makeatletter
\renewcommand\AB@affilsepx{\\\protect\Affilfont}
\makeatother
\affil[ ]{\small\tt\{mateo.obregon,r.l.hill\}@ed.ac.uk}
\makeatletter
\renewcommand\AB@affilsepx{ \protect\Affilfont}
\makeatother
\affil[ ]{\small\tt f.blain@wlv.ac.uk}
\affil[ ]{\small\tt\{m.fomicheva,l.specia\}@sheffield.ac.uk}
\affil[ ]{\small\tt lisa.yankovskaya@ut.ee}

\date{}

\begin{document}
\maketitle

\begin{abstract}
    Translating text into a language unknown to the text's author, dubbed outbound translation, is a modern need for which the user experience
    has significant room for improvement, beyond the basic machine translation facility. We demonstrate this by showing 
    three ways in which user confidence in the outbound translation, as well as its overall final quality, can be affected:
    backward translation, quality estimation (with alignment) and source paraphrasing.
    In this paper, we describe an experiment on outbound translation from English to Czech and Estonian. We examine the effects of each proposed feedback module and further focus on how the quality of machine translation systems influence these findings and the user perception of success.
    We show that backward translation feedback has a mixed effect on the whole process: it increases user confidence in the produced translation, but not the objective quality.
\end{abstract}

\section{Introduction}
\label{sec:intro}

When dealing with machine translation (MT) on the web, most of the attention of the research community is paid to \emph{inbound translation}.
In this scenario, the recipients are aware of the MT process, and thus it is their responsibility to interpret and understand the translated content correctly.
For an MT system, it is sufficient to achieve such quality that allows a recipient to get the gist of the meaning of texts on webpages.

For \emph{outbound translation}, it is the other way round: the responsibility to create the content in the way that it is correctly interpreted by a recipient lies on the authors of the message.
The main issue is that the target language might be entirely unknown to them.
Prototypically it is communication by email, filling in foreign language forms, or involving some other kind of interactive medium.
The focus in this scenario is placed not only on producing high-quality translations but also on reassuring the author that the MT output is correct.

One of the approaches to improving both quality and authors' confidence, first employed in this scenario by \citet{zouhar:ptakopet_pilot}, is to provide cues that indicate the quality of MT output as well as suggest possible rephrasing of the source.
They may include backward translation to the source language, highlighting of the potentially problematic parts of the input, or suggesting paraphrases.
Except for preliminary work by \citet{zouhar-novak:2020}, the impact of individual cues has not yet been properly explored.

In this paper, we present the results of a new experiment on outbound translation.
Building on the previous works, the focus was expanded to investigate the influence of different levels of performance of the underlying MT systems, as well as utilizing a much greater range and diversity of participants and evaluation methods.

Native English speakers were tasked to produce text either in Czech or in Estonian with an outbound translation system in an e-commerce context.
Every user also reported a confidence score upon finishing each stimulus trial. A native Czech or Estonian speaker later evaluated each final translation for fluency and adequacy. The set of available cues varied for each participant from stimuli to stimuli, following a controlled experimental design, in order to determine the impact of specific combinations of cues on the self-reported confidence and the final translation quality.

For our study, we made use of the \ptakopet{} system \citep{zouhar:enabling_outbound}.
This bespoke software was specifically developed to examine user behavior when testing machine translation user interfaces, especially in the context of outbound translation.\footnote{The code for this project and also the experiment data are available as open-source. \href{https://github.com/zouharvi/ptakopet}{github.com/zouharvi/ptakopet}}

\begin{figure*}[ht]
    \center
    \includegraphics[width=0.85\linewidth]{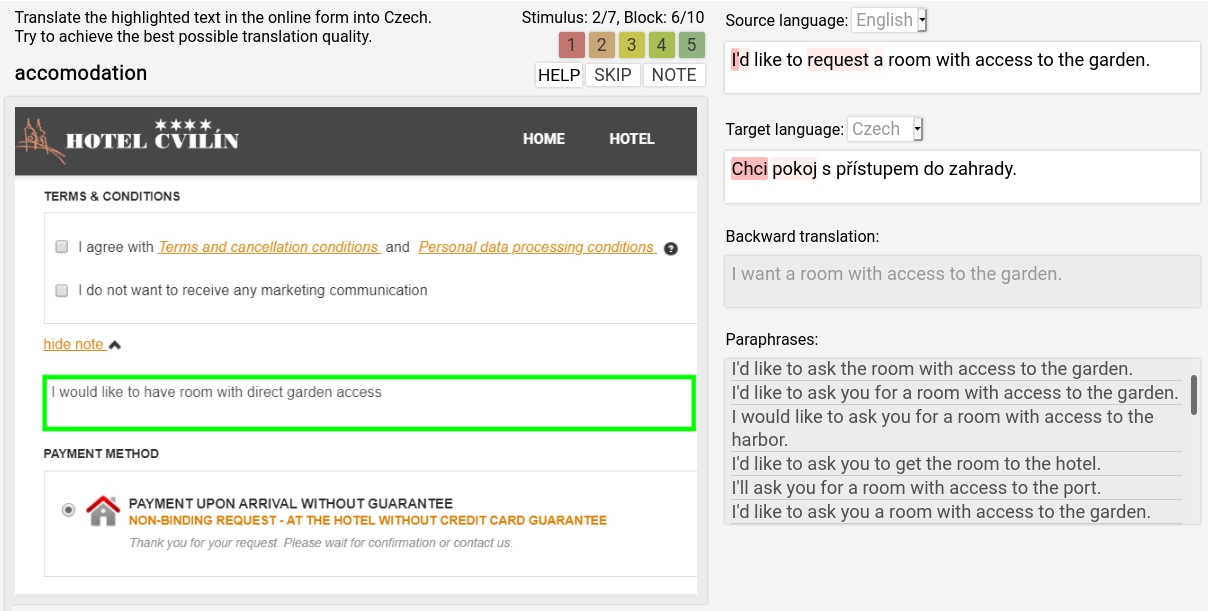}
    \caption{Screenshot of the \ptakopet{} experiment interface with all modules enabled. Only the two white text areas on the right are editable. The first is the source (from which forward translation is made) and the second is the final output (from which backtranslation is shown). Editing the second text area was purely optional as it was a language unknown to the participant.}
    \label{fig:ptakopet_screen}
    \vspace{-0.3cm}
\end{figure*}

The structure of the paper is as follows.
After an overview of the related work in \Cref{sec:relwork}, we present the environment for the outbound translation we used for the experiment, including the MT systems and modules that provided cues to the users, in \Cref{sec:ot-envir}.
\Cref{sec:data} describes the data that we collected during the experiment, and in \Cref{sec:res} we further analyze them to reveal and discuss various aspects of our approach to outbound translation.
We conclude with the main findings in \Cref{sec:concl}.

\section{Related Work}
\label{sec:relwork}

Despite recent advances in neural machine translation (NMT) quality, resulting in output comparable to human professionals in specific settings \citep{microsoft:chinese:parity:2018,popel:cubitt}, it is far from reasonable to blindly believe that the output of MT systems is perfectly accurate.
It should thus not be simply included in an email or another message without some means of verification. Feedback in this scenario is needed, which would tell users if the translation is correct and ideally even give instructions on how to improve it.

A related area of interactive machine translation (IMT) focuses mainly on either post-editor scenarios \citep{gomez-adaptation,Sanchis-Trilles-etal,underwood2014evaluating,alabau2016learning} or generally scenarios in which users are able to produce the translation themselves and the system only aims to speed it up or improve it \citep{santy-etal-2019-inmt}.

Outbound translation differs from common IMT scenarios by the fact that the user does not speak the target language, and hence operates on the MT result only in a limited way.

The first work to deal with this task by \citet{zouhar:ptakopet_pilot} focused on working with Czech-German MT in context of asking and reformulating questions.
A preliminary experiment on the effect of translation cues has been carried out by \citet{zouhar-novak:2020}, but it was conducted on a much smaller scale both in terms of participants and annotators and with non-native speakers of English.
This may have affected the results that differ in some aspects, especially in the usefulness of the word-level quality estimation.

\section{Environment for Testing Outbound Translation}
\label{sec:ot-envir}

In order to test the effect of different cues, we utilized \ptakopet{}, a web-based tool for outbound translation. The tool provides machine translation together with cues in the form of backward translation, quality estimation and paraphrasing. These cues are intended to help the user arrive at a better translation and increase their confidence in the produced output. The tool is modular, allowing the modules for MT and cues to be either replaced with others or turned on and off.

By linking a collection of sample stimuli to the tool it can also be used to conduct experiments.
Participants are asked to react to stimuli by formulating texts in a language known to them and producing and editing translations in a language they do not know.
The set of cues they are presented with may vary.
The users are also asked to report their confidence in the produced output.

In this experiment, each participant was presented with a sequence of scenes, interacting with the outbound translation system in each of them.
\Cref{fig:ptakopet_screen} shows an example of a scene and user interaction.
In the following sections, we describe the main components of the experiment.

\subsection{Stimuli} 
\label{sec:stimuli}

We used screenshots of web forms (real-world examples from the e-commerce domain) as stimuli.
Every screenshot displayed an excerpt of a web form containing a text field for open queries with a specific query already pre-filled and highlighted in a green rectangle.
For example, \Cref{fig:ptakopet_screen} shows a form at hotel webpages with a pre-filled special request.

This query, or rather its message, is what should be translated.
Apart from the query, the screenshot captured elements of the webpage that should make it easier and faster for the user to understand the intended message and its context.
The stimuli are also accompanied by a short description of the website's domain (e.g. \emph{accommodation}) above the screenshot for the same purpose.

The dataset consists of 70 screenshots and corresponding pre-filled queries in English.%
\footnote{%
As the pre-filled queries were conceived by non-native speakers of English, they may contain grammatical errors.
The intention behind them is always understandable, though.
}
It was selected from a collection of 462 such screenshots, collated by six annotators.%
\footnote{%
Available at \href{http://hdl.handle.net/11234/1-3622}{hdl.handle.net/11234/1-3622}.%
}
The annotators were instructed to look for web forms with text boxes that could be filled with text which would require translation.
We were not interested in fields such as names, addresses, numbers or pre-defined lists of values (e.g. countries).
We emphasized that the collection should consist of a broad variety of domains, but the particular choice of domains and websites was up to the annotators.

\subsection{Modules}
\label{sec:modules}

The set of available modules (backward translation \texttt{BT}, quality estimation \texttt{QE}, paraphrasing \texttt{PP}), as well as the choice of the MT system, was randomized for every user for every stimulus.
We denote a specific cue configuration by the modules present, e.g. \texttt{BT PP}.
\Cref{fig:ptakopet_screen} shows an example of modules' outputs, given a user's rephrasing of the query from the stimulus.

\paragraph{Machine Translation.}

We used three MT systems for Czech (differing in speed and training data size) and one for Estonian.
All of the systems were trained in both directions: the forward systems translate from English, whereas the opposite direction is used as a backward translation cue.
All the MT systems follow the Transformer model architecture \citep{vaswani2017transformer} design, though student systems make use of the simplified simple recurrent unit and other modifications described in \citet{germann-EtAl:2020:WMT}.
\Cref{tab:mt} shows how the MT systems performed in terms of BLEU score \citep{papineni2002bleu} on the test set of WMT18 News task \citep{bojar-etal-2018-findings}.

The \emph{Czech~3} system is the winning MT model of Czech--English News Translation in WMT 2019 \citep{popel-EtAl:2019:WMT}, having been trained on 58M authentic sentence pairs and 65M backtranslated monolingual sentences.%
\footnote{%
In the opposite direction, 48M monolingual sentences have been used to create synthetic data.%
}

The training proposed by \citet{germann-EtAl:2020:WMT} was used for a CPU-optimized student model \emph{Czech~2}.
It was created by the knowledge distillation \citep{kim2016sequence} method on translations generated by Czech~3.
Although it has been trained solely on synthetic data, its performance in the news domain falls behind the teacher only by 0.5 to 3.0 BLEU points, depending on the translation direction.
We included it mainly due to its speed as shown in \Cref{sec:data}.

The design of the \emph{Czech~1} system is identical to Czech~3.
The only difference is that the former was trained only on a subsample of 5M sentence pairs from CzEng 1.7 \citep{bojar2016czeng}.
This system was chosen to simulate performance on less resourceful language pairs.

The \emph{Estonian} system uses the same construction procedure as Czech~2. The teacher system utilized in knowledge distillation was internally trained for us by the authors of \citet{germann-EtAl:2020:WMT}.

\begin{table}[t]
    \centering
    \resizebox{0.6\columnwidth}{!}{
    \begin{tabular}{lcccc}
    \toprule
                & \emph{X}$\leftarrow$EN   & \emph{X}$\rightarrow$EN \\
    \midrule
    Czech~1     & 19.57     & 25.04 \\
    Czech~2     & 23.85     & 32.71 \\
    Czech~3     & 26.00     & 33.11 \\
    Estonian    & 25.85     & 31.61 \\
    \bottomrule
    \end{tabular}
    }
    \caption{Performance of utilized MT systems in BLEU score evaluated on WMT18 test set; higher is better.}
    \label{tab:mt}
    \vspace{-0.4cm}
\end{table}

\paragraph{Quality Estimation.}

QE is the task of predicting the quality of an MT output without relying on reference translation, as opposed to traditional evaluation based on automatic metrics (BLEU, TER, etc.). 
We have used QE to predict potential translation errors at the word-level which in turn, combined with a source-target token-level alignment algorithm,%
\footnote{%
It was provided by FastAlign \citep{dyer-etal-2013-simple} models trained on bitext from CzEng 2.0 \citep{DBLP:journals/corr/abs-2007-03006} and OPUS collection \citep{tiedemann-2012-parallel} for English-Czech and English-Estonian, respectively.
Measured on 10 queries sampled from the dataset of stimuli and their translations produced by the Czech~3 and Estonian systems, the F1 score of English tokens alignment exceeds 80\% in both cases.%
}
enables us to identify the source words that have led to those translation errors.
QE suggestions are presented by red word highlighting (see \Cref{fig:ptakopet_screen}).


\begin{table}[t]
    \centering
    \resizebox{0.85\columnwidth}{!}{
    \begin{tabular}{lccc}
    \toprule
                        & \multirow{2}{*}{EN$\rightarrow$ET} & \multicolumn{2}{c}{EN$\rightarrow$CS} \\
                        &       & Czech 1    & Czech 3   \\
    \midrule
    Accuracy            & 0.74  & 0.70  & 0.77 \\
    F1$_{\texttt{BAD}}$    & 0.37  & 0.32  & 0.14 \\
    F1$_{\texttt{OK}}$     & 0.83  & 0.81  & 0.87 \\
    MCC                 & 0.28  & 0.23  & 0.12 \\
    \bottomrule
    \end{tabular}
    }
    \caption{Performance of the word-level QE system on the outputs of our MT systems.}
    \label{tab:gb_qe_results}
    \vspace{-0.4cm}
\end{table}

We note that word-level error annotation is a hard and costly task.
Thus, available data for building systems to predict word-level errors is scarce.
To circumvent this issue we relied on a feature-based approach which exploited information from the neural MT system (i.e. a glass-box approach to QE) and did not require large amounts of data for training.
Glass-box features have been successfully used for QE of statistical MT \cite{blatz2004confidence,specia2013quest} and have been recently shown to be effective for sentence-level QE of neural MT systems \cite{fomicheva2020unsupervised}.
To accommodate for the different types of MT models used in this work, including a student model \emph{Czech~2}, we did not use the full set of features from \citet{fomicheva2020unsupervised} but instead relied on simple subset of log-probability based features:

\begin{itemize}[noitemsep,nolistsep,leftmargin=*]
    \item Log-probability of the word
    \item Log-prob. of the previous word
    \item Log-prob. of the next word
    \item Average log-prob. of the translated sentence
    \item Number of characters in the word
\end{itemize}

We build a binary gradient boosting classifier to predict word-level quality. To train the classifier we collected a small curated dataset with translation error annotation. 
Although the annotation is binary%
\footnote{%
In addition, each translated word labeled as \texttt{BAD} was manually annotated with a subcategory: minor, major or critical.
However, due to the heavy imbalance of the data, we did not use this fine-grained annotation to train the QE system.%
}
(\texttt{OK/BAD} class), the dataset is heavily imbalanced.
To alleviate this issue, we over-sampled the minority class (\texttt{BAD}). 



We randomly split the data for each MT system into train (80\%) and test (20\%). In addition to accuracy, we report F1 for each class and Matthews correlation coefficient (MCC) as proposed by \citet{fonseca-etal-2019-findings} for imbalanced data. \Cref{tab:gb_qe_results} shows these results for Estonian and Czech.

We observed that F1 for the \texttt{BAD} class is much lower than F1 for \texttt{OK}.
This indicates the difficulty of our QE models in correctly predicting the minority class.
The reasons for that are as follows. First, log-probabilities might not contain enough information to predict major or critical issues. In particular, critical issues concern the mistranslation of specific elements in the text (e.g. numbers or named entities), which is beyond the scope of the glass-box features used in our experiments. We plan to investigate other light-weight features that could better capture this information. Secondly, on average, MT quality is quite high (even for weaker models) and therefore, the vast majority of the words belong to the positive class.

\paragraph{Paraphraser.}

This module was expected to provide users with a potential rephrasing of their inputs from which they may draw inspiration for alternative translations. The paraphraser is based on pivoting, i.e. a round-trip translation via a pivot language.
\citet{federmann.etal19} showed that pivoting is an effective way of generating diverse paraphrases, especially if done via linguistically unrelated languages.
A larger set of pivot languages should further increase the diversity of paraphrases.

Our paraphrasing system performed two-step English-to-English translation through 41 pivot languages.
It is based on T2T-multi-big model from \citet{machacek-etal-2020-elitr}, a multi-lingual Transformer-big \citep{vaswani2017transformer} model with a shared encoder and decoder.
It has been trained on 231M sentence pairs sampled from the OPUS collection \citep{tiedemann-2012-parallel}.
Given a sentence, the model yielded 41 variants. In order not to overwhelm users, the paraphrases are then grouped so that two paraphrases with the same bag of words excluding stop words end up in the same group. In the end, users are presented with a list of one random representative from each group, sorted by the group size in descending order. The paraphrases suggested by multiple languages should thus appear at the top. To achieve reasonable response time (ca. 3s), the service has been run on a GPU.

\Cref{tab:paraph} shows the performance of the paraphraser in terms of BLEU score, evaluated on a subset of the Quora Question Pairs dataset.%
\footnote{%
\href{https://www.quora.com/q/quoradata/First-Quora-Dataset-Release-Question-Pairs}{quoradata.quora.com/First-Quora-Dataset-Release-Question-Pairs}%
}
The subset consists of 4000 question pairs, with 2000 pairs containing real paraphrases, and 2000 containing similar sentences with a different meaning. The two cases are respectively denoted by $+$ and $-$.
The produced outputs seem to be more similar to real paraphrases than to fake ones, which corresponds to what we observed for source sentences with twice as high BLEU scores.

\begin{table}[t]
\centering
    \resizebox{0.8\columnwidth}{!}{
    \begin{tabular}{lccc}
    \toprule
               & Source   & Ref$+$    & Ref$-$ \\
    \midrule
    Paraphrase   & 35.46 & 13.59     & 7.42 \\
    Source       & --    & 29.04     & 15.71 \\
    \bottomrule
    \end{tabular}
    }
    \caption{
    Performance of the paraphraser in BLEU averaged across all languages. The produced paraphrase is compared either with the source sentence, or with the reference, which can be a real paraphrase ($+$), or a similar sentence with a different meaning ($-$). For illustration, a comparison of the source sentence with the two types of reference is added.
    }
    \label{tab:paraph}
    \vspace{-0.2cm}
\end{table}

\subsection{Self-reported confidence}

Users were asked to submit their rephrased English query and its translation by reporting their confidence in the produced translation.
They specified how much they trusted the translation on a standard Likert scale from 1 (least) to 5 (most).

\section{Data Collected in the Experiment}
\label{sec:data}

During a single scene, the participant saw a stimulus, worked on it and then finished it either by rating their confidence or by describing the reason for skipping.
The participant was continuously presented with the translation output and the cues.
We logged all incoming data as well as requests to the modules and their responses together with timestamps.

In total, $52$ English speaking participants joined our experiment, out of whom $49$ were native speakers of English.
There were $70$ scenes, each with a unique stimulus, prepared for every participant.
After filtering out the scenes which we found invalid as they contained either no input from the users or were not finished, the total number of scenes to be analyzed was $2486$.
The participants thus succeeded in completing $48$ scenes on average.
As shown in \Cref{tab:time_summary}, the distribution of completed scenes over different configurations appears to be balanced.


Since one of the goals of \ptakopet{} is to facilitate work with MT, we also focused on the time participants had to spent in the interface together with the number of their actions%
\footnote{%
We measure actions by the number of forward translation requests because they are present in every configuration.%
}
needed to finish stimuli.
They are summarized in \Cref{tab:time_summary}.
It is clear that the short response times of student models (Czech MT 2 and Estonian) encourage the users to perform more actions, while still spending less time on one scene on average.

\begin{table}[t]
    \centering
    \resizebox{0.8\columnwidth}{!}{
    \begin{tabular}{lcccc}
        \hline
        Config. & \# Scenes & Time [s] & Actions & Pace [s] \\ \toprule
Czech 1 & 610 & 59 & 1.85 & 44 \\
Czech 2 & 643 & 42  & 3.66 & 17 \\
Czech 3 & 601 & 52 & 1.66 & 41 \\
Estonian  & 632  & 46  & 3.06 & 22 \\ \midrule
BT QE PP & 307 & 49 & 2.77 & 27 \\
BT QE & 331 & 47 & 2.70 & 25 \\
BT PP & 304 & 51 & 2.42 & 31 \\
QE PP & 298 & 55 & 3.11 & 27 \\
BT & 311 & 50 & 2.31 & 31 \\
QE & 304 & 46 & 2.84 & 25 \\
PP & 302 & 49 & 2.68 & 28 \\
- & 285 & 46 & 2.08 & 34 \\
\midrule
Total & 2486 & 49 & 2.62 & 28 \\






%
 \bottomrule
    \end{tabular}
    }
    \caption{Summary of collected scenes, median time, mean number of actions, and median pace (time per one action) aggregated over all scenes across different configurations. Time and pace are in seconds; actions are computed from translation requests made. For the two variables involving time, median was used instead of mean in order to avoid the effect of outlier scenes where the user was inactive for a longer time period.
	}
    \label{tab:time_summary}
    \vspace{-0.25cm}
\end{table}

\section{Evaluation and Results}
\label{sec:res}

Having recorded the essential interactions of participants with Ptakopět, we further analyzed the collected data, especially user inputs and their translations.

\paragraph{Viable inputs.}
Unless a participant skipped a scene, it was concluded by confirming the \emph{final input} and its translation.
We were also interested in examining intermediate complete sentences which users considered and later abandoned.
We call these \emph{viable intermediate inputs}.
The collection of such inputs was possible because the Ptakopět tool continually records user's interaction.
We set the minimum time without any edit for an input to be sent to the forward translation module to 1000 ms.
Despite this relatively long period, still many incomplete or erroneous inputs were recorded, perhaps while the user was deliberating.
We thus used a simple heuristic to extract the viable ones.


For an input to be considered viable, it had to end with a full stop, an exclamation mark, or the same token as the final input ended.
Furthermore, its length had to be within a 25\% margin around the length of the final input without whitespaces.%
\footnote{%
This rule discredits inputs meant to be viable, where the very last token was later edited, though.
Manual examination of the data verified the efficacy of the heuristic.%
}



Whereas each confirmed scene by design resulted in $1$ final input and its translation, the number of intermediate viable inputs (non-final) was $0.62$.
Their average length was $98.43\%$ of the final input.

\begin{table}[t]
    \centering
    \resizebox{\columnwidth}{!}{
    \begin{tabular}{lcccccc}
        \toprule
        Config &
        \makecell{SRC\\ STI} &
        \makecell{TGT\\ SRC} &
        \makecell{TGT\\ STI} &
        \hspace{-0.2cm}Fluency\hspace{-0.2cm}&
        \hspace{-0.2cm}Overall\hspace{-0.2cm}&
        \hspace{-0.2cm}Conf.\hspace{-0.2cm} \\
        \toprule
        Czech 1 & 4.46 & 4.38 & 4.02 & 4.22 & 4.10 & 3.40 \\
Czech 2 & 4.47 & 4.48 & 4.14 & 4.23 & 4.19 & 3.73 \\
Czech 3 & 4.47 & 4.63 & 4.26 & 4.45 & 4.33 & 3.70 \\
Estonian & 4.58 & 4.31 & 4.05 & 4.28 & 4.14 & 3.51 \\
\midrule
BT QE PP & 4.48 & 4.46 & 4.11 & 4.31 & 4.19 & 3.81\threeS \\
BT QE & 4.49 & 4.51\oneS & 4.18 & 4.33 & 4.26 & 3.71\threeS \\
BT PP & 4.52 & 4.45 & 4.16 & 4.29 & 4.22 & 4.07\threeS \\
QE PP & 4.43\oneS & 4.42 & 4.03 & 4.29 & 4.12 & 3.41\threeS \\
BT & 4.54 & 4.50 & 4.20 & 4.30 & 4.24 & 4.15\threeS \\
QE & 4.46 & 4.39 & 4.05 & 4.26 & 4.13 & 2.84 \\
PP & 4.50 & 4.43 & 4.09 & 4.28 & 4.17 & 3.61\threeS \\
-- & 4.50 & 4.43 & 4.09 & 4.28 & 4.17 & 3.61 \\
\midrule
Total & 4.49 & 4.45 & 4.12 & 4.29 & 4.19 & 3.59 \\

        \bottomrule
    \end{tabular}
    }
    \caption{%
        Average quality of final inputs and their translations, and average self-reported confidence of participants across various configurations. We mark configurations of cue combinations if they are significantly different from the configuration with no cues according to Mann-Whitney~U test ($p < 0.05$\oneSmark{}, $0.001$\threeSmark{}).
    }
    \label{tab:final_res}
    \vspace{-0.2cm}
\end{table}

\paragraph{Evaluation of translation quality.}
The extracted viable inputs and their translations were rated for quality and adequacy by 12 Czech and 3 Estonian native speakers.
For each viable input, the annotators were shown the source, its translation and the corresponding stimulus.
They were asked to rate on the scale from 1 (least) to 5 (most) the following statements:

\begin{itemize}[noitemsep,nolistsep,leftmargin=*]
\item \textit{SRC-STI:} The meaning of user input corresponds to what is entered in the form shown in the image.
\item \textit{TGT-SRC:} The meaning of the translation corresponds to the user input.
\item \textit{TGT-STI:} The meaning of the translation corresponds to what is entered in the form shown in the image.
\item \textit{Fluency:} The translation is fluent (including typography, punctuation, etc.)
\item \textit{Overall: } The overall translation quality including both adequacy with respect to the stimulus and fluency is high.
\end{itemize}

On average, we collected $7.15$ assessments per viable input.
The inter-rater agreement measured by Kripendorff's alpha was 0.47 and 0.48 for Czech and Estonian, respectively.

\begin{figure}[t]
    \center
    \includegraphics[width=\linewidth]{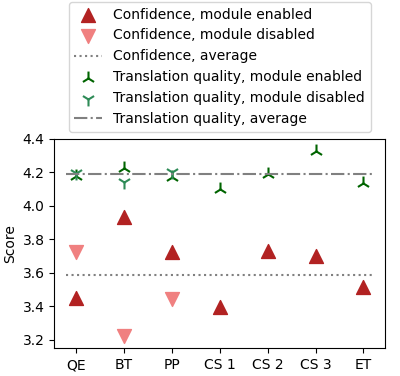}
    \caption{Effect of different MT systems and the presence and absence of every module on self-reported user confidence and translation quality.}
    \label{fig:modules_mts}
    \vspace{-0.2cm}
\end{figure}

\paragraph{Data Normalization.}

Because of data imbalance in favor of high confidence, we normalized the self-reported user confidences using the following formula: $x' = \frac{x-min}{max-min} \times 4 + 1$. The $min$ and $max$ values were taken individually for every participant. This only affected those who never used $1$ or $2$ in their self-reported confidences. We did not apply this normalization to the quality annotations, because the annotators used the whole scale in almost all cases. The overall average of all confidence judgments decreased from $3.72$ to $3.59$ by this normalization.

This only helped with the imbalance a little. To avoid strong assumptions about the underlying process, we did not normalize the data to have zero mean and standard deviation of 1 for every feature dimension. This would also have made any interpretation less intuitive.

\paragraph{Results on final inputs.}

\begin{figure}[t]
    \center
    \includegraphics[width=1.0\linewidth]{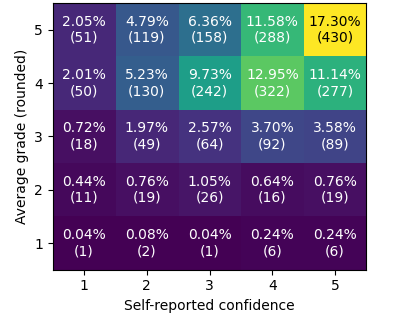}
    \caption{Distribution (percentage and absolute count) of quality annotation (rounded overall) and self-reported user confidence.}
    \label{fig:rating_heatmap}
    \vspace{-0.3cm}
\end{figure}

\Cref{tab:final_res} shows the average evaluation scores of final confirmed inputs, accompanied by average self-confidence scores across various configurations. For clarity, we illustrate the same results in \Cref{fig:modules_mts}. Comparing the Czech MT systems, their ranking with respect to the Overall score corresponds to the results of the automatic evaluation in the news domain shown in \Cref{tab:mt}.

Interestingly enough, Czech 2 received an average confidence score comparable to its teacher model Czech 3 (see in \Cref{fig:modules_mts}).
The results of comparison across different combinations of cues suggest that configurations with backtranslation feedback enabled achieved better performance in terms of the overall quality.
In such cases, the users also felt more confident. Unlike for overall quality, the effect of an available backward translation cue on user confidence was statistically significant by Mann-Whitney U test for $0.6$ point difference ($U=24243.5, p < 0.0001$).

Conversely, quality estimation cues appear not to be useful, which the users also noted.
Unfortunately, the presence of paraphrases increased user confidence, but decreased the objective translation quality.
These results are in contrast with the work of \citet{zouhar-novak:2020}.
We attribute this difference to an insufficient number of samples and also a more homogeneous composition of participants (all foreign PhD students studying in the Czech Republic) in their work.

Note that users who had knowledge of some other Slavic language (Polish or Russian) on average expressed higher confidence ($3.95$) and also produced translations of higher quality ($4.44$).
The effects of different modules on their work were closer to the effects described in \citet{zouhar-novak:2020}.

As seen in \Cref{fig:rating_heatmap}, a significant proportion of the scenes (\textasciitilde{}\hspace{0.02cm}41\%) received 4 or 5 on both self-reported confidence and overall translation quality. Although these high scores are positive in terms of industry progress, it makes the quality-confidence dependency harder to analyze.

\begin{table}[t]
    \centering
    \resizebox{\columnwidth}{!}{
    \begin{tabular}{lccccc}
        \toprule
        &
        \makecell{TGT\\ SRC} &
        \makecell{TGT\\ STI} &
        \hspace{-0.2cm}Fluency\hspace{-0.2cm}&
        \hspace{-0.2cm}Overall\hspace{-0.2cm}&
        \hspace{-0.2cm}Conf.\hspace{-0.2cm}\\
        \toprule
        
SRC-STI & 0.07 & 0.64 & 0.23 & 0.50 & 0.08 \\
TGT-SRC &              & 0.69 & 0.68 & 0.71 & 0.14 \\
TGT-STI &              &              & 0.67 & 0.88 & 0.15 \\
Fluency &              &              &              & 0.84 & 0.13 \\
Overall &              &              &              &              & 0.14 \\

        \bottomrule
    \end{tabular}
    }
    \caption{
        Correlation between all quality annotations variables and self-reported user confidence.
    }
    \label{tab:corr_heatmap}
    \vspace{-0.2cm}
\end{table}

\Cref{tab:corr_heatmap} shows expected rating behavior in terms of correlations.
We can see that Fluency is mostly correlated with TGT-SRC and TGT-STI adequacies and less with SRC-STI adequacy, which should affect the translation fluency only slightly.%
\footnote{%
In a scenario where the SRC-STI adequacy is lowered by typos in Source, which then also negatively affects the translation process and also the Fluency.%
}
We also see that TGT-STI adequacy and Fluency affects the Overall rating the most, which accords with its definition.
Self-reported user confidence correlates the least with all the rest, but slightly more with TGT-STI, TGT-SRC and Overall scores, which we consider positive.

\begin{figure*}[t]
    \center
    \includegraphics[width=0.6\linewidth]{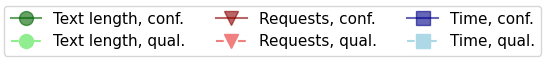}
    \includegraphics[width=0.95\linewidth]{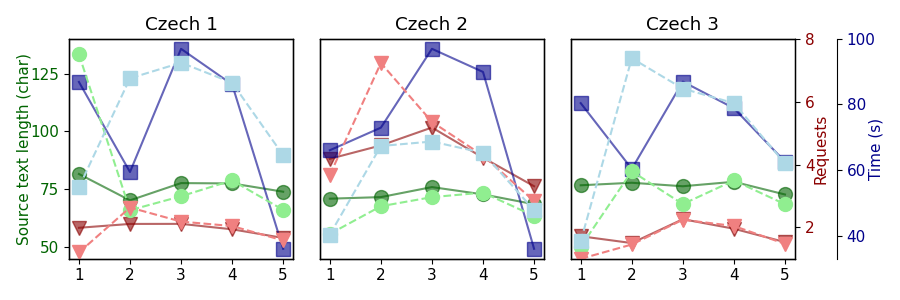}
    \includegraphics[width=0.5\linewidth]{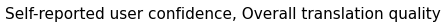}
    \caption{Relationship between the scores (self-reported user confidence in darker colors and  overall translation quality in lighter colors) and average scene time, forward translation request count and input length in characters (y-axes). Best viewed in color.}
    \label{fig:exp1}
    \vspace{-0.3cm}
\end{figure*}

\paragraph{MT comparison in detail.}

\Cref{fig:exp1} shows the average spent time per stimulus as well as the number of forward translation requests and input length in characters with respect to the confidence and overall translation quality for submitted translations. The figure is split into three graphs, each corresponding to one of the Czech MT systems. 

Input text length does not appear to affect the overall translation quality significantly, while it seems to affect users' self-reported confidence. 

The curves for time spent, although different in absolute values, peak in the middle (rating 3) and have the lowest values for scores of 1 and 5.
This may happen because the stimulus was either easy to complete, or the users did not work on this stimulus diligently.
It is supported by the fact that they did not report low confidences in these instances.
A similar trend, although less pronounced, can be seen with the number of requests.

We can also notice that the Czech 2 system has the lowest times despite also having a vastly higher number of executed requests.
The request delay was the same for all MT systems, so in this case, the users recognized that they did not have to wait so long for getting a translation back and hence sent more requests.
This is one of the possible explanations for why in \Cref{fig:modules_mts} the average self-reported confidence for this system is on par with its teacher model, Czech 3, despite being less performant objectively.

The degree of interactivity appears to be the main factor affecting these MT systems profiles. The figures of Czech 1 and Czech 3 look very similar even though they vary greatly in performance and only have their speeds in common (slower than Czech 2).

\paragraph{Intermediate vs. final.}

Having also intermediate viable inputs at our disposal, we explored how quality changes in the transition from intermediate to final inputs.
We excluded those scenes that contain no viable intermediate input, which accounts for almost 69\%.

Although our heuristics can filter out most of the intermediate inputs which are not viable, some of those remaining can be still considered defective.
They may contain a typo, artifacts of unfinished rephrasing or may miss important information. 
These non-viable inputs must be excluded from the comparison, as the user would unlikely submit them or they could be easily fixed by a spell-checker.
We manually examined all intermediate viables and excluded the defective ones from the following statistics.

\begin{table}[t]
    \centering
    \resizebox{\columnwidth}{!}{
    \begin{tabular}{lccccc}
        \toprule
        Config &
        \makecell{SRC\\ STI} &
        \makecell{TGT\\ SRC} &
        \makecell{TGT\\ STI} &
        \hspace{-0.2cm}Fluency\hspace{-0.2cm}&
        \hspace{-0.2cm}Overall\hspace{-0.2cm}\\
        \toprule
        BT QE PP & \textcolor{darkred}{-0.19}\threeS & \textcolor{darkgreen}{+0.10} & \textcolor{darkgreen}{+0.04} & \textcolor{darkgreen}{+0.05} & \textcolor{darkgreen}{+0.08} \\
BT QE & \textcolor{darkred}{-0.14}\threeS & \textcolor{darkgreen}{+0.16}\twoS & \textcolor{darkgreen}{+0.03} & \textcolor{darkgreen}{+0.04} & \textcolor{darkgreen}{+0.03} \\
BT PP & \textcolor{darkred}{-0.12} & \textcolor{darkgreen}{+0.14} & \textcolor{darkgreen}{+0.16} & \textcolor{darkgreen}{+0.12} & \textcolor{darkgreen}{+0.17} \\
QE PP & \textcolor{darkred}{-0.20}\threeS & \textcolor{darkgreen}{+0.03} & \textcolor{darkred}{-0.13}\oneS & \textcolor{darkgreen}{+0.02} & \textcolor{darkred}{-0.07}\oneS \\
BT & \textcolor{darkred}{-0.24}\threeS & \textcolor{darkgreen}{+0.33}\threeS & \textcolor{darkgreen}{+0.10} & \textcolor{darkgreen}{+0.10} & \textcolor{darkgreen}{+0.11} \\
QE & \textcolor{darkred}{-0.11}\oneS & \textcolor{darkred}{-0.01} & \textcolor{darkred}{-0.10} & \textcolor{darkred}{-0.05} & \textcolor{darkred}{-0.04} \\
PP & \textcolor{darkred}{-0.11}\twoS & \textcolor{darkgreen}{+0.09} & \textcolor{darkred}{-0.05} & \textcolor{darkred}{-0.00} & \textcolor{darkred}{-0.04} \\
-- & \textcolor{darkred}{-0.02} & \textcolor{darkgreen}{+0.16} & \textcolor{darkgreen}{+0.16} & \textcolor{darkgreen}{+0.04} & \textcolor{darkgreen}{+0.06} \\
\midrule
Total & \textcolor{darkred}{-0.15}\threeS & \textcolor{darkgreen}{+0.11}\threeS & \textcolor{darkgreen}{+0.01} & \textcolor{darkgreen}{+0.04} & \textcolor{darkgreen}{+0.03} \\

        \bottomrule
    \end{tabular}
    }
    \caption{%
        Average difference of quality between intermediate viable and final inputs and their translations for all combinations of available cue modules. Statistical significance was calculated by Wilcoxon signed-rank test ($p < 0.05$\oneSmark{}, $0.01$\twoSmark{}, $0.001$\threeSmark{})
    }
    \label{tab:viable_final}
    \vspace{-0.4cm}
\end{table}

\begin{table*}[t]
    \centering
    \resizebox{0.8\textwidth}{!}{
    \begin{tabular}{ll}
        \toprule
        Inter   & \begin{small}I teach my son \textbf{English} with the 'Learning Time with Timmy' series on Youtube.\end{small} \\[-0.7ex]
        & \begin{small}Učím svého syna \textbf{Angličana} \emph{/Englishman/} se seriálem „Learning Time with Timmy“ na Youtube.\end{small} \\[-0.7ex]
Final   & \begin{small}I teach my son \textbf{English language} with the series 'Learning Time with Timmy' series on Youtube.\end{small} \\[-0.7ex]
        & \begin{small}Učím svého syna \textbf{anglický jazyk} se seriálem „Learning Time with Timmy“ na Youtube.\end{small} \\[-0.7ex]
\midrule
Inter   & \begin{small}Why was I not able to make a payment \textbf{by} mobile?\end{small} \\[-0.7ex]
        & \begin{small}Proč jsem nemohl zaplatit \textbf{za} \emph{/for/} mobil?\end{small} \\[-0.7ex]
Final   & \begin{small}Why was I not able to make a payment \textbf{from} my mobile?\end{small} \\[-0.7ex]
        & \begin{small}Proč jsem nemohl zaplatit \textbf{z} mobilu?\end{small} \\[-0.7ex]
\midrule
Inter   & \begin{small}What documents do I need to have if my ID \textbf{has expired}?\end{small} \\[-0.7ex]
        & \begin{small}Jaké doklady potřebuji, když mi \textbf{vypršel} průkaz totožnosti?\end{small} \\[-0.7ex]
Final   & \begin{small}What documents do I need to have if my ID is \textbf{out of date}?\end{small} \\[-0.7ex]
        & \begin{small}Jaké dokumenty potřebuji, když je můj průkaz \textbf{zastaral} \emph{/got obsolete/}?\end{small} \\[-0.7ex]

        \bottomrule
    \end{tabular}
    }
    \caption{%
        Examples of user interaction with the Ptakopět system.
        In the top two, the rephrasing of the intermediate input resulted in an improved final translation, in the bottom one the final translation worsened.
    }
    \label{tab:examples}
    \vspace{-0.3cm}
\end{table*}

\Cref{tab:viable_final} shows the average difference in the quality of intermediate and corresponding final inputs and their translations.
The greatest improvement in the Overall score is again achieved by configurations utilizing backtranslation feedback, although the difference is not statistically significant.
What is significant, though, are some TGT-SRC scores including the BT configuration.
It shows that the translation of the final input is on average more adequate to the source than the translation of the intermediate inputs.
Nevertheless, the effect on the TGT-STI adequacy is marginal due to negative differences in the SRC-STI adequacy score.
These can be justified by the fact that any modification of the original query in the stimulus might have been considered as a shift in meaning by the annotators, although in reality the original intention could be still understandable.

In \Cref{tab:examples}, we show three examples of the intermediate and the final inputs with their translations to Czech.
In the top two, the rephrasing helped to improve the translation quality: (1) by adding a word ``language'' to prevent translating ``English'' as a Czech word for ``Englishmen'', or (2) by substituting a preposition.
Conversely, the replacement of the verb ``has expired'' by a phrase ``out of date'' led to a drop in translation quality.
This is due to a grammatical error and use of the Czech expression meaning ``got obsolete'', which indeed sounds old-fashioned in this context.


\vspace{-0.1cm}
\section{Conclusion}
\label{sec:concl}

In this paper, we demonstrated through an experiment the effect of three translation cues on user confidence and translation quality.

The backward translation cue proves to be a powerful means to enhance user confidence in MT.
At the same time, it neither increase nor decrease significantly the translation quality.
The fact that backtranslation feedback has a marginal effect to objective quality but greatly increases user confidence is surprising because it is the most intuitive low-effort approach to outbound translation scenarios which can be done even with publicly available MT systems.

The paraphraser seems to increase user confidence less (compared to not being present), with no or slightly negative impact on the translation quality.
Without a better method to generate diverse and still adequate paraphrases, employing this cue is questionable.
The effect of word-level quality estimation appears to be even more questionable.
We attribute it mainly to the underlying word-level models, which may not be mature enough for user-facing applications.

Despite the loss in objective translation quality, the CPU-optimized student MT model either managed to maintain its teacher's high trustworthiness or compensated for it by its speed.

\paragraph{Future work.}
Scores in both user confidence and overall translation quality annotation cluster together.
Having the distribution less concentrated by changing the underlying task with stimuli or by working with more low resource languages could reveal stronger dependencies between individual variables.

We limited ourselves to only three baseline solutions to help in outbound translation.
In the future work, inspiration could be drawn from the approaches of interactive machine translation systems and these could be adapted for the purposes of outbound translation.


\section*{Acknowledgments}

Sincere thanks to Chris Burns and the three anonymous reviewers for their thorough review and helpful comments.

This project has received funding from the grants H2020-ICT-2018-2-825303 (Bergamot) of the European Union and 19-26934X (NEUREM3) of the Czech Science Foundation. 
The work has also been supported by the Ministry of Education, Youth and Sports of the Czech Republic, Project No. LM2018101 LINDAT/CLARIAH-CZ.

\section*{Ethics}
All participants were recruited online and had to complete an informed consent form using a secure Qualtrics survey before they could progress to taking part in the experiment. Data was anonymized, recorded and stored in accordance with ACM protocol. Ethical clearance was confirmed by the School of Informatics Ethics Committee at the University of Edinburgh (Reference RT 4058). Participants were offered a £20 Amazon voucher as compensation for their time upon completion of the experiment.

\bibliography{mybib}
\bibliographystyle{acl_natbib}

\end{document}